\documentclass[runningheads]{llncs}

\usepackage{eccv}

\usepackage{picinpar}

\usepackage{mathrsfs}

\usepackage{algorithm}
\usepackage{algorithmic}
\usepackage{multirow}

\usepackage{makecell}
\usepackage{color}
\usepackage{stfloats}
\usepackage{float}    
\usepackage{graphicx}

\usepackage{eccvabbrv}

\usepackage{graphicx}
\usepackage{booktabs}

\usepackage[accsupp]{axessibility}  % Improves PDF readability for those with disabilities.

\usepackage[pagebackref,breaklinks,colorlinks,citecolor=eccvblue]{hyperref}
\usepackage{orcidlink}

\begin{document}

\title{LogoSticker: Inserting Logos into Diffusion Models for Customized Generation} 

\titlerunning{LogoSticker}

\author{Mingkang Zhu\inst{1} \quad
Xi Chen\inst{2} \quad
Zhongdao Wang\inst{3} \\
Hengshuang Zhao\inst{2} \quad
Jiaya Jia\inst{1,4}}

\authorrunning{M. Zhu et al.}

%\institute{	The Chinese University of Hong Kong \and
%The University of Hong Kong \and
%Noah's Ark Lab \and SmartMore}
%\institute{$^{1}$The Chinese University of Hong Kong \quad $^{2}$The University of Hong Kong \\ $^{3}$Noah's Ark Lab \quad $^{4}$SmartMore}
\institute{$^{1}$CUHK \quad $^{2}$HKU  \quad $^{3}$Huawei Noah's Ark Lab \quad $^{4}$SmartMore}

\maketitle

\begin{abstract}

Recent advances in text-to-image model customization have underscored the importance of integrating new concepts with a few examples. Yet, these progresses are largely confined to widely recognized subjects, which can be learned with relative ease through models' adequate shared prior knowledge. In contrast, logos, characterized by unique patterns and textual elements, are hard to establish shared knowledge within diffusion models, thus presenting a unique challenge. To bridge this gap, we introduce the task of logo insertion. Our goal is to insert logo identities into diffusion models and enable their seamless synthesis in varied contexts. We present a novel two-phase pipeline LogoSticker to tackle this task. First, we propose the actor-critic relation pre-training algorithm, which addresses the nontrivial gaps in models' understanding of the potential spatial positioning of logos and interactions with other objects. Second, we propose a decoupled identity learning algorithm, which enables precise localization and identity extraction of logos. LogoSticker can generate logos accurately and harmoniously in diverse contexts. We comprehensively validate the effectiveness of LogoSticker over customization methods and large models such as DALLE~3. \href{https://mingkangz.github.io/logosticker}{Project page}.


\end{abstract}

\section{Introduction}
\label{sec:intro}

\begin{figure*}

  \centering
\includegraphics[width=\linewidth]{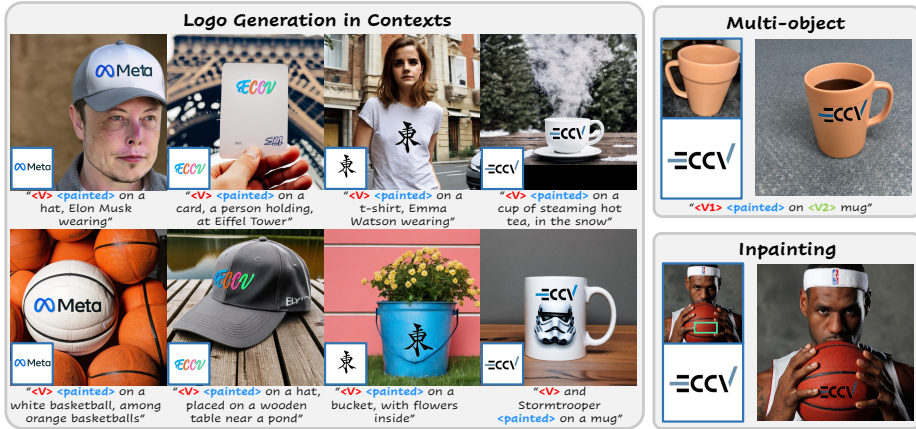}

     \captionof{figure}{Given a logo with fine-grained details, our method LogoSticker enables accurate distilling of its identity to diffusion models,  thus supporting coherent text-to-image generation in diverse scenarios. It can also be extended to multi-object customization, and logos inpainting on user-given images.}
     \label{fig1}

\end{figure*}

Recent advancements in text-to-image generative models have demonstrated unprecedented capabilities in generating high-quality images conditioned on text prompts \cite{latent_diffusion, imagen, dalle2, dalle3, nichol2022glide, sauer2023stylegant,kang2023scaling,gafni2022makeascene}. To allow the generation of user-specified subjects, various customization methods have been proposed to integrate user-provided concepts into these models \cite{gal2022textual, Ruiz_2023_CVPR,kumari2022customdiffusion,avrahami2023break,chen2023anydoor}. Despite their success on  everyday objects, these methods still have  difficulties inserting concepts without enough commonly shared knowledge, like logos or texts. For example, Stable Diffusion \cite{latent_diffusion} has adequate prior knowledge of pets. Even without customization, it can generate Corgis similar in appearance to a user-specified Corgi using only text prompts. However, it is nontrivial to synthesize precise English text using prompts \cite{chen2023textdiffuser}, not to mention logos, which can additionally contain  unique patterns and non-English texts. 

Logos, which have extremely diversified shapes and appearances, are typically difficult to have commonalities but are crucial in applications like marketing and advertising. Therefore, in this work, we aim to tackle the challenge of logo insertion: given a user-provided logo to the diffusion model, our goal is to enable the model to recognize the logo, learn its identity accurately, and enable coherent generations of the logo in various scenarios.

Without adequate shared prior knowledge of diversified logos, the task of logo insertion is nontrivial. The first challenge is that diffusion models are not fully proficient in accurately conveying relationships \cite{huang2023reversion}, including the relation of painting things on different objects, which is essential for our task. This deficiency is more pronounced for logos, as logos are diverse and complicated and the models can hardly find references in their knowledge about where the diverse logos should appear, and how would they interact with other objects. The second challenge is the ambiguity in concept extraction from training images: It is usually not obvious which concept should be extracted from the training images. Previous methods generally rely on the diffusion model's adequate prior knowledge of commonplace items (e.g., dog or cat) \cite{gal2022textual,Ruiz_2023_CVPR}, so that the class name token can attend the object they intend to learn. This approach, however, is not viable for learning diverse and complicated logos. Usual relevant class name tokens for logos generally cannot accurately attend diverse logos we aim to learn, as shown in Fig. \ref{attn_1}. Without this ability, precise logo identity extraction is unlikely.

We propose LogoSticker to tackle this task. To address the first challenge, we need to encapsulate the relation of painting things on various objects into diffusion models. There have been attempts to represent an object-to-object relation by optimizing a text token on a few images \cite{huang2023reversion}. However, we find that: the relation cannot be effectively learned and generalized with limited training data; the difficulties of learning the relation for different objects are different. These observations lead us to introduce the Actor-Critic Relation Pre-training strategy. We curate a diverse relation dataset of various objects to paint things on. Then we fine-tune a text token and the text encoder to learn the relation. Since learning to paint things on some objects is significantly more difficult than others, we employ an actor-critic strategy to adjust the sampling probabilities of different objects. Specifically, we use a CLIP model \cite{clip} as the critic to evaluate whether the model has learned to paint things on certain objects. More samplings are given to objects that have not been mastered. In this way, the learned relation becomes stronger and more balanced among different objects.

To tackle the second challenge, we decouple the logo identity learning process into two parts: binding the target logo to a text token, and accurately learning the logo identity. Specifically, we propose to generate two distinct sets of training images for this purpose: the logo token binding set and the logo identity learning set.  We generate the logo token binding set by pasting the logo on random locations of solid color backgrounds, and generate the logo identity learning set by pasting the logo on random locations of natural scenes. To make the model recognize the logo, we fix the U-Net parameters and optimize a special token on the logo token binding set so that the special text token is bound with the target logo and can attend the logo precisely. Then we fine-tune the U-Net on the logo identity learning set with the optimized special token in the prompt to accurately extract the logo identity and map it to the special text~token.

With the aforementioned methods, LogoSticker can stick diverse logos in different contexts, as demonstrated in Fig. \ref{fig1}. We quantitatively and qualitatively evaluate LogoSticker against other customization methods. We also conduct a user study to demonstrate  human evaluators' preference for our method's generations. Comparisons with state-of-the-art models including DALLE~3 \cite{dalle3} and text generation models like Textdiffuser-2 \cite{textdiffuser2} and AnyText \cite{anytext} further demonstrate LogoSticker's advantage in contextualizing logos precisely and harmoniously. We also demonstrate its versatility by showing several applications.

Our contributions can be summarized threefold:
(1) We study a new problem of logo insertion, which aims to internalize user-provided logos accurately and empower consistent and diverse contextualizations. Existing customization methods largely focus on widely recognized concepts by utilizing the model's sufficient prior knowledge, while we make the exploration into complex and diverse logos. (2) We propose a novel two-phase pipeline LogoSticker for this task, consisting of the actor-critic relation pre-training algorithm followed by the decoupled identity learning algorithm. Our pipeline enables precise and seamless generations of logos in various contexts. (3) Experiment results demonstrate the effectiveness of LogoSticker %our method 
over state-of-the-art customization methods and large models including DALLE~3 in logo contextualization. 

\section{Related Work}
\label{sec:related_works}

\textbf{Text-to-image generative models.} Significant progress has been made in text-to-image generative models in recent years. Taking advantage of the development of diffusion models \cite{NEURIPS2020_4c5bcfec, song2021denoising,song2021scorebased,sohldickstein2015deep} and large-scale cross-modal models like CLIP \cite{clip} and T5 \cite{T5},  synthesizing high-fidelity and diverse images based on text descriptions has become a reality. State-of-the-art large-scale text-to-image diffusion models such as Stable Diffusion \cite{latent_diffusion}, Imagen \cite{imagen}, DALLE~2 \cite{dalle2}, and DALLE~3 \cite{dalle3} have significantly advanced these capabilities. ControlNet \cite{controlnet} introduces more conditions to improve diffusion models' controllability. However, we find that it is still nontrivial for them to follow detailed text descriptions and precisely reconstruct user-provided logos. Our method focuses on integrating user-provided logos into these models so that they can be elicited using text prompts. The objective is to facilitate the coherent generation of these logos across varied contexts while maintaining their distinct characteristics.

\noindent \textbf{Customization of generative models.} Customized image synthesis focuses on implanting user-provided subjects into diffusion models and facilitating their generation in different contexts. Textual Inversion \cite{gal2022textual} tries to reconstruct user-provided subjects by optimizing a text token and adding it to the text encoder's dictionary. Dreambooth \cite{Ruiz_2023_CVPR} uses an existing rare token and fine-tunes the model weights. Recent developments have explored customizing multiple user-specific concepts simultaneously, either by finetuning \cite{avrahami2023break, kumari2022customdiffusion,liu2023cones} or by zero-shot methods \cite{chen2023anydoor,sarukkai2023collage}. ReVersion \cite{huang2023reversion}, on the other hand, tries to optimize a text token representing a general object-to-object relation. However, these methods typically focus on customizing subjects or concepts that generative models have sufficient prior knowledge of. Their performance on logos remains unsatisfactory. In contrast, our method aims to introduce complex and diverse logos, of which models have little prior knowledge, to the output domain of text-to-image diffusion models and enable their coherent generation in various contexts.

\noindent \textbf{Visual text generation.}
Although recent text-to-image generative models can create vibrant and complex images, they often fall short of generating legible and cohesive texts. Some works claim this drawback is due to that the CLIP embedding does not contain character-level information of words in prompts which could be crucial for generating readable texts \cite{dalle2}. Therefore, recent methods like Imagen \cite{imagen} utilize large language models like T5 \cite{T5} to achieve better text rendering quality. GlyphControl \cite{yang2023glyphcontrol},  Textdiffusers \cite{chen2023textdiffuser, textdiffuser2}, and AnyText \cite{anytext} incorporate the glyph information to better condition the text generation. These methods mainly focus on English texts and cannot customize text generation. It is also nontrivial for them to follow precise descriptions of the properties or styles of the desired texts. Our method can customize text generation, which enables diffusion models to follow precise specifications of desired text properties. Our pipeline can also deal with non-English texts effortlessly.

\section{Methods}

\begin{figure}[t]
  \centering
  \includegraphics[width=0.9\textwidth]{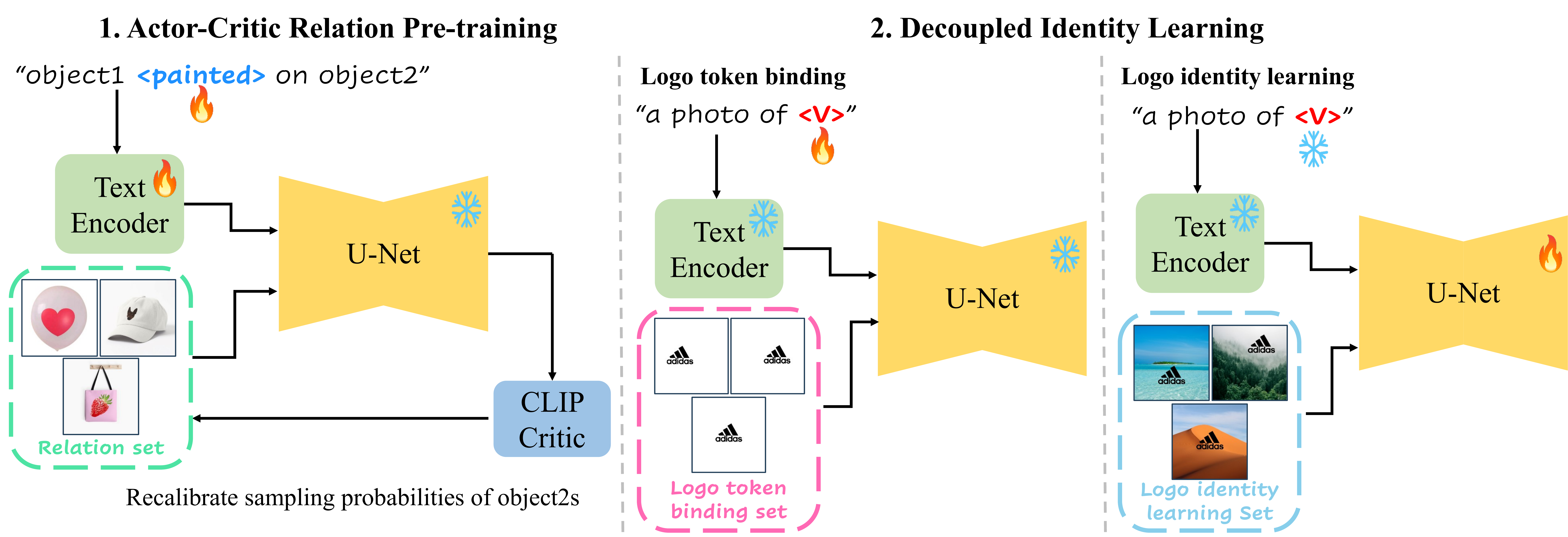}

  \caption{\textbf{The overall pipeline of our proposed LogoSticker.} (1) We first pre-train the text encoder and a token {$<$painted$>$} in an actor-critic fashion to learn the relation of logo placement in various contexts effectively. (2) We build the logo token binding set and optimize another special token {$<$V$>$} to bind it with the target logo so that the target logo in training images can be localized. Then, we build a more complex logo identity learning set and fine-tune the U-Net to capture the logo identity precisely.}
  \label{fig:method}

\end{figure}

Given a user-provided logo, we aim to generate the logo in different contexts precisely and cohesively. Specifically, we want the generated logo to retain its unique identity while being positioned correctly and angled suitably across diverse settings. We achieve this via a two-phase pipeline LogoSticker,  in which each phase incorporates the relation of appearing in various contexts for logo placement and precisely learns the logo identity correspondingly. The overall pipeline is presented in Fig. \ref{fig:method}. In this section, we first lay the groundwork by providing an overview of text-to-image diffusion models in Subsection \ref{Preliminaries}. In Subsection \ref{Relation_Learning}, we present how we map the relation of appearing in various contexts to a text token, which in turn enables coherent logo generation in various scenes. Subsequently, we demonstrate how to make the diffusion model accurately recognize and internalize the identity of the logo in Subsection \ref{Decoupled_Identity_Learning}.

\subsection{ Preliminary}
\label{Preliminaries}

Diffusion models are a class of generative models that learn data distributions through progressively denoising Gaussian distribution samples \cite{song2021denoising,song2021scorebased, NEURIPS2020_4c5bcfec}. Contemporary diffusion models are often executed in the compressed latent space of an autoencoder for efficiency \cite{latent_diffusion,dalle3}. They use a pre-trained autoencoder $\mathcal{E}$ to map images $x$ into their latent counterparts $z=\mathcal{E}(x)$. Then the latent diffusion model is trained to denoise the latent code $z$.  The diffusion model can be conditioned on a vector $c = \gamma_\theta(P)$, where $\gamma_\theta$ is a pre-trained text encoder and $P$ is the conditioning prompt. Then the latent diffusion loss can be cast as:
\begin{equation}
\begin{aligned}
L_{LDM}(\theta) = \mathbb{E}_{z,P,t,\epsilon \sim \mathcal{N}(0,\,1)} \left[ ||\epsilon-\epsilon_\theta(z_t, t, \gamma_\theta(P))||_2^2 \right],
\end{aligned}
\end{equation}
where $t$ is timestep, $z_t$ is noised latent at time $t$, and $\epsilon_\theta$ is the denoising U-Net \cite{ronneberger2015unet}. Our method 
is built on top of Stable Diffusion \cite{latent_diffusion}, a widely recognized and publicly accessible latent diffusion model trained on the LAION dataset \cite{laion}.

\subsection{Actor-Critic Relation Pre-training}
\label{Relation_Learning}
As we aim for contextually appropriate and accurate logo generations in various contexts, we want the diffusion models to be capable of painting logos coherently on various objects. However, diffusion models sometimes even fall short in accurately conveying this relation for common items. Thus, attempts like ReVersion \cite{huang2023reversion} have been made to encapsulate a general object-to-object relation into a special text token, including the relation of ``painted on''. This is closely related to our purpose. However, upon a detailed examination of ReVersion, we find the ``painted on'' relation it learns is relatively weak and observe several limitations of this approach: (1) The relation of ``painted on'' cannot be effectively learned and generalized with too few and similar exemplars as in \cite{huang2023reversion}. (2) The difficulties of learning the relation of ``painted on'' different objects vary a lot. Therefore, this method might be less effective for logos, since logos have complex details and models lack prior knowledge of logos including the knowledge of where the logos should appear, and how would they interact with other objects in the real world. Hence, we propose the following strategies to tackle these problems and make the learned relation stronger.

\noindent \textbf{Relation data collection.} We first collect a diverse dataset of the relation ``object1 painted on object2''. ``Object2s'' are subjects that we intend to paint the logo on, such as shirts, hats, and mugs. Instead of real logos, we choose ``object1s'' to be commonplace objects that the diffusion model possesses more prior knowledge of, like dogs, apples, or stars. We want the relation can be learned more easily and precisely, leveraging the model's encapsulated understandings of how the ``object1s'' are interacting with these ``object2s''. The relation dataset consists of 20 different ``object2s'' and each ``object2'' has 3 to 4 training images of different poses to ensure diversity. With this diverse dataset, we want the relation of ``painted on object2s'' to generalize for logos.

\noindent \textbf{Actor-critic sampling.} Different objects have different shapes, textures, and rarities. It is natural that painting something onto different objects has varying levels of difficulty. For example, painting patterns on a shirt or a mug is more common and easier than painting patterns on an egg or a slice of toast. Therefore, we need some mechanism to ensure that we can paint the logo more equally onto various ``object2s''.  
To this end, we propose to utilize a pre-trained CLIP model \cite{clip} as a critic to evaluate the diffusion model's effectiveness in applying common patterns onto different objects, throughout the relation learning process. With some ad-hoc ``object1s'', we denote the prompt ``object1 painted on object2'' as $c_{obj2}$ and the prompt using special text token $<$painted$>$  ``object1 $<$painted$>$ on object2'' as $c_{obj2<painted>}$. Then we denote the CLIP score of generating ``object1'' onto ``object2'' as $s_{obj2}$ in the following \cref{clip_score}:

\begin{equation}
   \label{clip_score}    
      \begin{aligned}
         s_{obj2} = CLIP(LDM(c_{obj2<painted>}), c_{obj2}).
  \end{aligned}       
\end{equation}

Then the Actor-Critic Relation Pre-training algorithm can be summarized in Algorithm \ref{Alg1}, which adjusts the sampling probabilities of training images of ``object2s'' every $f$ iterations. It uses the CLIP critic to rate whether the model has understood how to paint on each ``object2''. More frequent sampling will be applied on ``object2s'' that are not mastered, and vice versa. Besides optimizing a single text token, we also fine-tune the text encoder to strengthen the relation. 

\begin{algorithm}[ht]
\caption{Actor-Critic Relation Pre-training}\label{Alg1}
\begin{algorithmic}[1]
\REQUIRE 
Diffusion model $LDM$; Relation text token $<$painted$>$, CLIP critic model $CLIP$; constant $\lambda$; number of ``object2s'' $N$; probability recalibration frequency $f$; Number of training iterations $A$.\\
\ENSURE 
Pre-trained Diffusion Model.\\
    \FOR{$a$ = 1 to $A$}
       \STATE Fine-tune $LDM$ and $<$painted$>$ on relation dataset with sampling probability $p(obj2)$ for each ``object2'';
        \IF{ $a$ $\mod$ $f$ = 0 }
            \STATE calculate $s_{obj2}$ for each  $obj2$ by %Equation 
            \cref{clip_score};
            \STATE $\Bar{S} = \frac{1}{N} \sum_{obj_2} s_{obj2}$;
            \FOR{each  $obj2$ }
                \STATE $w(obj2) = \lambda^{\Bar{S} - s_{obj2}}$;
                \STATE $p(obj2) = \frac{w(obj2) }{\sum_{obj_2} w(obj2)}$;
            \ENDFOR
        \ENDIF
    \ENDFOR	
\end{algorithmic}
\end{algorithm}

\subsection{Decoupled Identity Learning}
\label{Decoupled_Identity_Learning}

\begin{figure}[t]
\centering
  \includegraphics[width=1.0\textwidth]{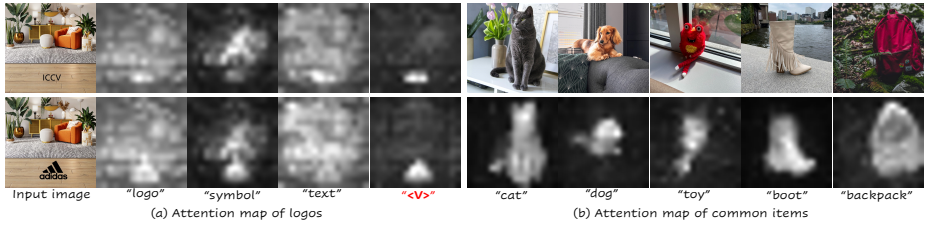}

  \caption{\textbf{Visualizations of attention maps} for: (a) Common synonyms of the word ``logo'' and our special token optimized on the logo token binding set.   (b) Class name tokens of commonplace items in Dreambooth's~\cite{Ruiz_2023_CVPR} dataset. Attention maps are computed by averaging attention activation across time steps and layers.}
  \label{attn_1}

\end{figure}

The Actor-Critic Relation Pre-training can encapsulate a stronger and more balanced ``painted on'' relation into diffusion models. Then our next task is to implant logo identities into the output domain of diffusion models. This task is rather challenging due to the model's lack of prior knowledge of diverse and complex logos. Without enough prior knowledge, it is difficult for the model to know which parts of the training images contain the logo to be learned. This easily results in unwanted concepts getting learned and intricate details of the logo not getting captured. We substantiate this issue by showing that usual relevant class names like ``logo'', ``symbol'', and ``text'' elicit no significant attention on diverse logo patterns in training  images. In Fig. \ref{attn_1}, we provide a visualization of attention maps on common items and logos, using Null-text Inversion \cite{mokady2022nulltext}. From Fig. \ref{attn_1}(b) we can see the attention maps of common items' class name tokens are very accurate. While from Fig. \ref{attn_1}(a), we can see that the word ``logo'' and its synonyms cannot precisely attend the logo regions at all. This challenge stems from the inherent complexity of logos, characterized by their diverse compositions, appearances, patterns, layouts, and textual elements. Consequently, the diffusion model lacks adequate and high-quality shared prior knowledge of logos, which is essential for effective logo recognition. This can make the learning process ambiguous and prone to overfitting to unwanted parts of training images. Thus the fine-grained details of logos cannot be preserved. Hence, to alleviate this issue, we decouple logo identity learning into  two parts and propose specific training data generation methods correspondingly.

\noindent \textbf{Logo token binding.} Since  the diffusion model cannot recognize complex logos, the first and foremost task is to make it able to identify the logo to be learned in images. Due to  the model's inability to localize logos in training images, fine-tuning all weights like Dreambooth \cite{Ruiz_2023_CVPR} would cause spurious correlations \cite{joshi2023spuco} or irrelevant concepts got learned. Therefore, we propose to constrain the learning scope and only optimize a special token $<$V$>$ using Textual Inversion \cite{gal2022textual}. Textual Inversion only allows the change of one single text token and thus does not learn much irrelevant information from training images. We want to bind the target logo to $<$V$>$ so that the logo can be accurately localized in training images. Textual Inversion generally collects training images that present the target concept in diverse settings, including different backgrounds and poses \cite{gal2022textual}. However, we observe in the case of diverse and complex logos, this setting easily leads to confusion about what concept should be learned. Therefore, we further propose to construct a distinct and less complex logo token binding set by pasting the logo onto random locations of solid color backgrounds contrasting to the color of the logo. We then use Textual Inversion to optimize a text token $<$V$>$ using our constructed logo token binding set. Although not able to reconstruct the logo, we find the optimized token $<$V$>$ can elicit significant attention on the logo region of training images, which facilitates further fine-grained logo identity extraction, as shown in Fig. \ref{attn_1}(a).

\noindent \textbf{Logo identity learning.} With the optimized special token $<$V$>$ accurately recognizing the target logo in images, our next task is to learn the logo identity accurately. Since the text token is tightly bound with the target logo, we can build a more complicated logo identity learning set. Specifically, we paste the logo onto random locations of various natural scenes with contrasting colors of the logo. Then, with the text token $<$V$>$ in the prompt, we finetune the weight of the U-Net to distill the logo identity into the special text token precisely.

\section{Experiments}
\label{sec:Experiments}

In this section, we illustrate our experiment setup, including the dataset we construct, evaluation metrics, and baseline customization methods. Then we compare our method LogoSticker with baseline customization methods quantitatively and qualitatively. Moreover, we present our conducted user studies, comparisons with large text-to-image models DALLE~3 \cite{dalle3}, and ControlNet \cite{controlnet}. We also compare LogoSticker with text generation models Textdiffuser-2 \cite{textdiffuser2}, and AnyText \cite{anytext}. Subsequently, we explore some extensions of our method, demonstrating its versatility. Finally, we present the ablation studies.

\subsection{Experimental Setup}

\textbf{Dataset.} We collect a dataset of 20 unique logos including texts with patterns, English texts, and Chinese texts with intricate calligraphy.\footnote{Existing customization methods generally collect dozens of subjects for evaluation. For example, Dreambooth \cite{Ruiz_2023_CVPR} compiles a dataset of 30 subjects while ReVersion \cite{huang2023reversion} assembles a dataset of 10 subjects.
} We test the generation of each learned logo identity in 20 different contexts. For quantitative analysis, each logo-context combination is synthesized using 5 different seeds to ensure comprehensive evaluation.

\noindent \textbf{Baselines.} To evaluate the effectiveness of our proposed LogoSticker, we compare it with 3 state-of-the-art baselines: Dreambooth \cite{Ruiz_2023_CVPR}, Textual Inversion \cite{gal2022textual}, and Dreambooth + ReVersion \cite{huang2023reversion}. We use the official implementation for ReVersion \cite{huang2023reversion}, and use the codes from diffusers \cite{diffusers} for Dreambooth and Textual Inversion. For baseline methods, we train them on the logo identity learning set, which consists of logos pasted onto random locations of natural images.

\noindent \textbf{Evaluation metrics.} Following prior practices \cite{Ruiz_2023_CVPR,gal2022textual}, we evaluate our LogoSticker and the baselines using two metrics: prompt fidelity and identity fidelity. Prompt fidelity measures the similarities between text descriptions and the generated images. Identity fidelity measures whether the generated images preserve the logo identity. Prompt fidelity, denoted as CLIP-T, is measured using the cosine distance between the CLIP \cite{clip} embeddings of text prompts and generated images. When evaluating prompt fidelity, we replace the special token {$<$V$>$} in the prompt with the word ``logo''. The identity fidelity, denoted as CLIP-I and DINO, is measured by CLIP \cite{clip} and DINO \cite{dino} scores correspondingly.

\begin{table}[t]
    \centering
      \caption{\textbf{Quantitative results and user study.} (Left) Quantitative metric comparisons of identity fidelity (DINO, CLIP-I) and prompt fidelity (CLIP-T). (Right) User study on  prompt fidelities (User PF) and logo fidelities (User LF).} 
    \label{tab1}  
  \resizebox{0.8\textwidth}{!}{
    \begin{tabular}{l c c c | c c}
        \hline
        Method & DINO $\uparrow$ & CLIP-I $\uparrow$ & CLIP-T $\uparrow$ & User PF $\uparrow$ & User LF $\uparrow$\\
        \hline
        Textual Inversion \cite{gal2022textual} &   0.228 & 0.648& 0.303  & 2.44 & 1.46\\
        DreamBooth \cite{Ruiz_2023_CVPR} & 0.227 & 0.739 & 0.278 & 3.01 & 3.21\\
        DreamBooth + ReVersion \cite{huang2023reversion}& 0.232 & 0.721 & 0.271 & 3.03 & 2.79\\
        LogoSticker (Ours) & 0.229 & 0.761 & 0.289&  4.37  & 4.46 \\
        \hline
    \end{tabular}  
  }  
\end{table}

\subsection{Comparisons}

\noindent \textbf{Quantitative results.} Tab. \ref{tab1} gives quantitative comparisons of identity fidelity  and prompt fidelity between our LogoSticker and baselines. From Tab. \ref{tab1}, we can see that LogoSticker outperforms all other methods in preserving logo identity. Existing methods generally have difficulties in learning and maintaining the logo identity. Intriguingly, we find that unlike in previous customization tasks \cite{Ruiz_2023_CVPR}, where the DINO score is usually close to the CLIP-I score, there exists a significant discrepancy between the DINO score and CLIP-I score for logos. We hypothesize that it might be because the DINO \cite{dino} is not a multi-modal model, and thus it is insensitive to textual elements and patterns such as logos. On the other hand, CLIP \cite{clip}, as a multi-modal model, can recognize texts and patterns quite well. CLIP has also been found to possess some OCR abilities and thus be susceptible to Typographic Attacks \cite{goh2021multimodal}, which aligns with our guess. We also conduct a user study to further evaluate all methods. The results are presented in Tab.~\ref{tab1}, from which we can see that users  generally prefer our  LogoSticker over others. This further verifies our method's superiority.

\begin{figure}[t]
  \centering
  \includegraphics[width=0.95\textwidth]{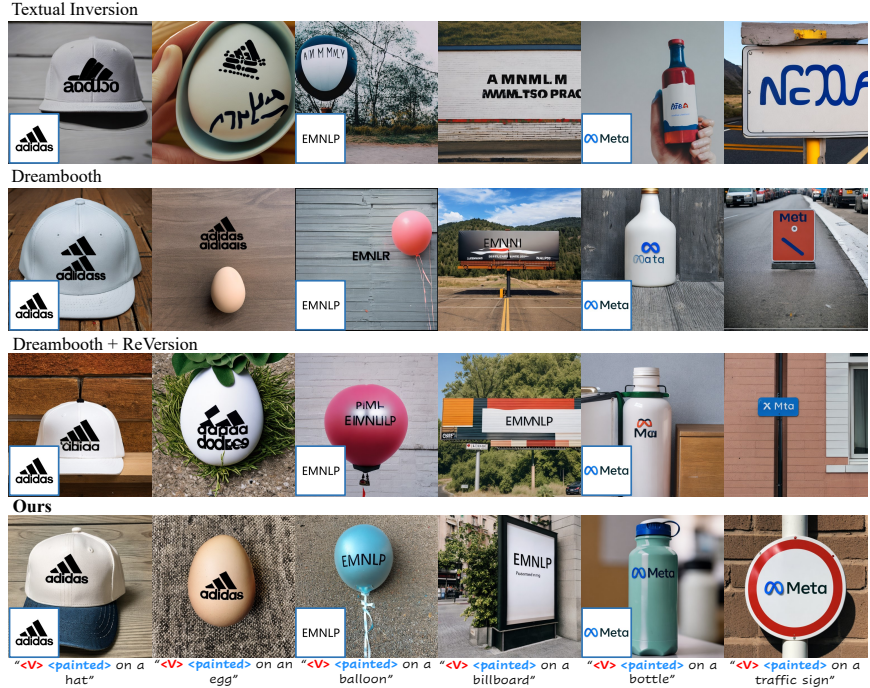}

  \caption{\textbf{Qualitative comparisons with baseline customization methods}. LogoSticker successfully preserves the logo identity while others struggle. LogoSticker %method 
  can synthesize the logo coherently on various objects. The logo identity is maintained even on curved objects or under various viewing positions.}  \label{qualitative_compare}

\end{figure}

\noindent \textbf{Qualitative results.}  We present the qualitative comparison of 6 logo-context pairs in Fig. \ref{qualitative_compare}. From the figure, we can see that Textual Inversion \cite{gal2022textual} cannot preserve the logo identity. Dreambooth \cite{Ruiz_2023_CVPR} and Dreambooth + ReVersion \cite{huang2023reversion} can preserve the logo identity better, but the logo identity is corrupted when it is generated onto other objects. Instead, our LogoSticker can preserve the logo identity excellently. Also, our generated logos are able to maintain their identities even on curved objects, inclined planes, or under different view angles.

\begin{figure}[t]

  \centering
  \includegraphics[width=1.0\textwidth]{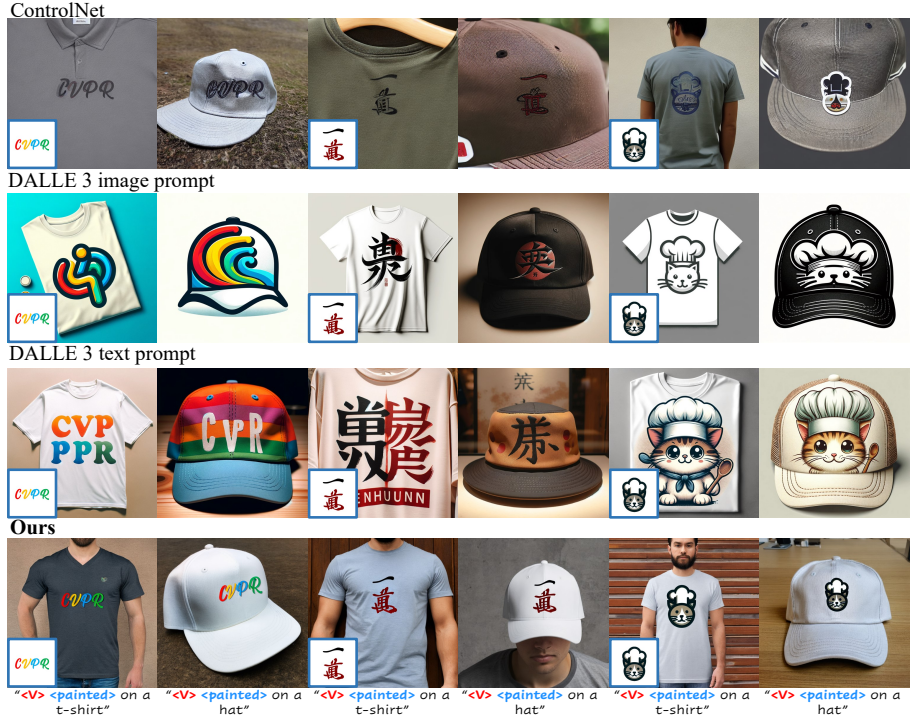}

  \caption{\textbf{Qualitative comparisons with large text-to-image models,} including ControlNet \cite{controlnet} and DALLE 3 \cite{dalle3} using both detailed text and image prompts.}
  \label{fig:compare gpt}

\end{figure}

\noindent \textbf{Comparisons with large text-to-image models.} We further compare our LogoSticker with ControlNet \cite{controlnet} and  GPT 4's \cite{openai2023gpt4} DALLE~3 \cite{dalle3}  using both detailed text prompts and uploaded image exemplars to describe target logos. From Fig. \ref{fig:compare gpt}, we can see that ControlNet can retain the overall shapes  of the logos to some level. However, both the details and colors of the logos are corrupted. Also, since ControlNet uses ad-hoc logo positions,  the coherence of synthesized images is usually compromised. For DALLE~3 with an image exemplar to describe the logo, the synthesized images can only preserve the high-level ideas of the uploaded logos, while losing all details. For DALLE~3 with detailed text prompts, we can see that it is also not able to preserve the intricate details of the logos using only textual descriptions. Moreover, we can see that DALLE~3 does not have the ability to synthesize legible Chinese texts, and the detailed specification of texts' colors also seems unlikely. In contrast, although built on Stable Diffusion \cite{latent_diffusion}, LogoSticker enables coherent generations of logos while accurately maintaining their identities. 

\begin{figure*}[t]

  \centering
  \includegraphics[width=1.0\textwidth]{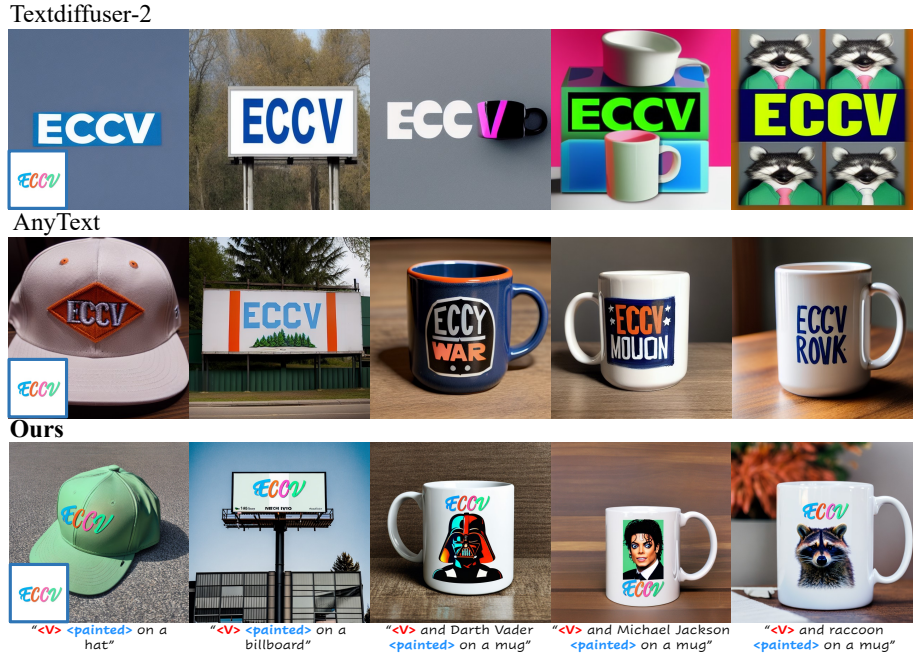}

  \caption{\textbf{Qualitative comparisons with text generation models,} including Textdiffuser-2 \cite{textdiffuser2} and AnyText \cite{anytext}.}
  \label{fig:text_comp}

\end{figure*}

\noindent \textbf{Comparisons with text generation models.}  We also compare LogoSticker with state-of-the-art text generation models Textdiffuser-2 \cite{textdiffuser2} and AnyText \cite{anytext} using detailed text prompts describing the target text. As demonstrated in Fig. \ref{fig:text_comp}, we can see both Textdiffuser-2 and AnyText fail to maintain the fine-grained details of the texts. Neither of them is able to follow the per-character text color specifications and the overall handwritten  style. Also, they cannot generate texts together with other concepts like ``Darth Vader'' or ``Michael Jackson''. Textdiffuser-2 \cite{textdiffuser2} even has difficulties generating texts onto objects like hats or mugs. In contrast, LogoSticker  
is free from these issues. It can accurately retain the per-character text color specifications and the overall writing  style of the text. The texts can also be painted onto objects with other concepts seamlessly.\vskip -0.2cm

\begin{figure}[t]
  \centering
  \includegraphics[width=1.0\textwidth]{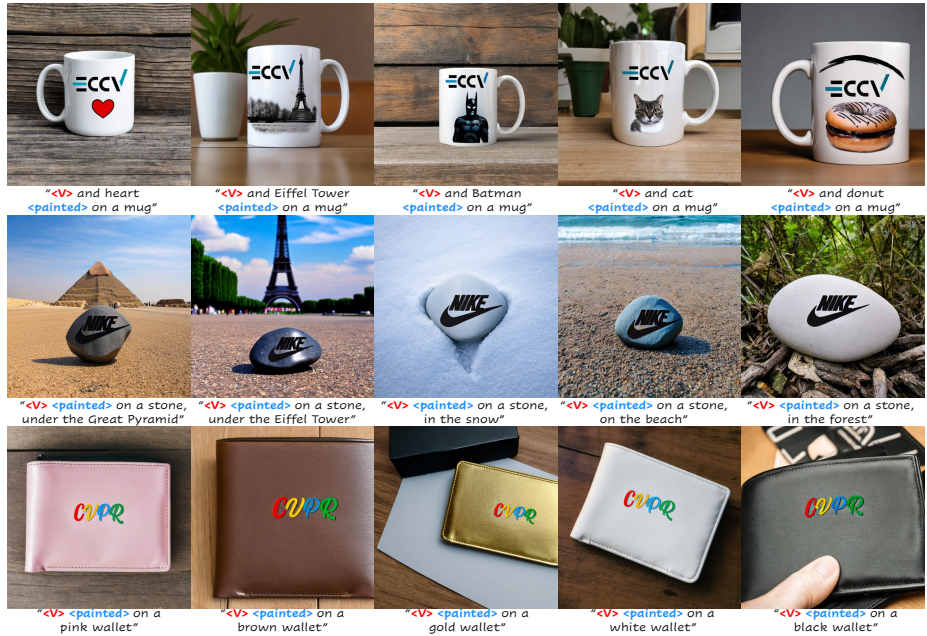}  

  \caption{\textbf{Application of LogoSticker} on more challenging contextualizations.}
  \label{fig:hard_case}

\end{figure}

\begin{figure*}[ht!]

  \centering
  \includegraphics[width=1.0\textwidth]{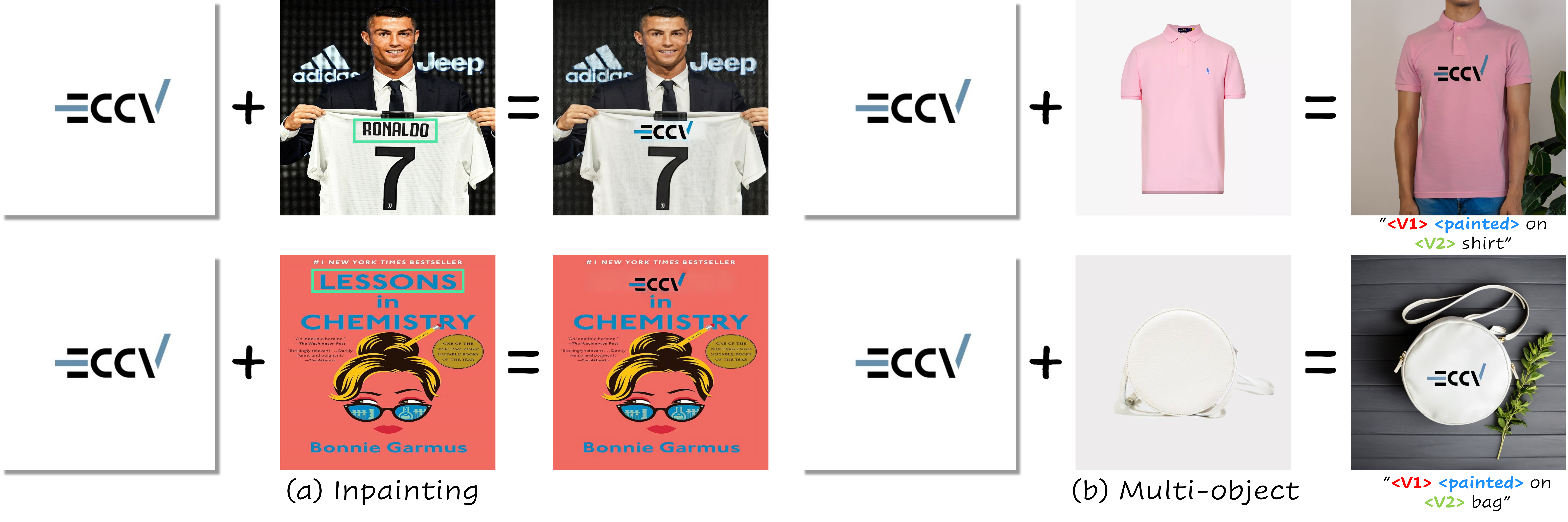}

  \caption{\textbf{Application of LogoSticker} on:
  (a) Inpainting. (b) Customizing both the logo and the context.}
  \label{fig:inpaint}
  
\end{figure*}

\subsection{More Applications}

\noindent \textbf{More challenging contextualizations.} We present more visualizations of our  LogoSticker's generations of logos in more diverse and challenging contexts.  In Fig. \ref{fig:hard_case}, we showcase 3 challenging cases to further validate the strength of LogoSticker. Row 1 of Fig. \ref{fig:hard_case} demonstrates that LogoSticker can paint the customized logo together with other logos onto objects coherently without any additional layout guidance. There are no overlaps or identity entanglements of both logos and identities of both logos are well preserved. Row 2 demonstrates that altering the overall scene in the inference prompt can still result in coherent and identity-preserving generations, while row 3 exemplifies that we can modify the properties of objects being painted on without hurting the logo's fidelity. These observations demonstrate that coherent and visually pleasing synthesis can be obtained with identity-preserved logos using LogoSticker.

\noindent \textbf{Inpainting.} LogoSticker adapts well with inpainting  \cite{latent_diffusion}. We can directly plug  it into existing inpainting pipelines \cite{latent_diffusion}. In this way,  LogoSticker is able to inpaint logos on user-provided images. From Fig. \ref{fig:inpaint}(a), we can see that the logo identity is well-preserved while the background identity is also maintained. 

\noindent \textbf{Multi-concept customization.} We conduct a preliminary experiment on  LogoSticker's compatibility with multi-concept customization. We simply integrate additional training images of the context into the training set during the logo identity learning phase. From Fig. \ref{fig:inpaint}(b), we can see that the identities of the polo shirt and round bag are well-preserved and the logo is painted on them seamlessly. This indicates the possibility of combining LogoSticker with advanced multi-concept customization methods for more complex customization tasks.

\subsection{Ablation Studies}

\begin{figwindow}[0,r,{
\includegraphics[width=0.4\textwidth]{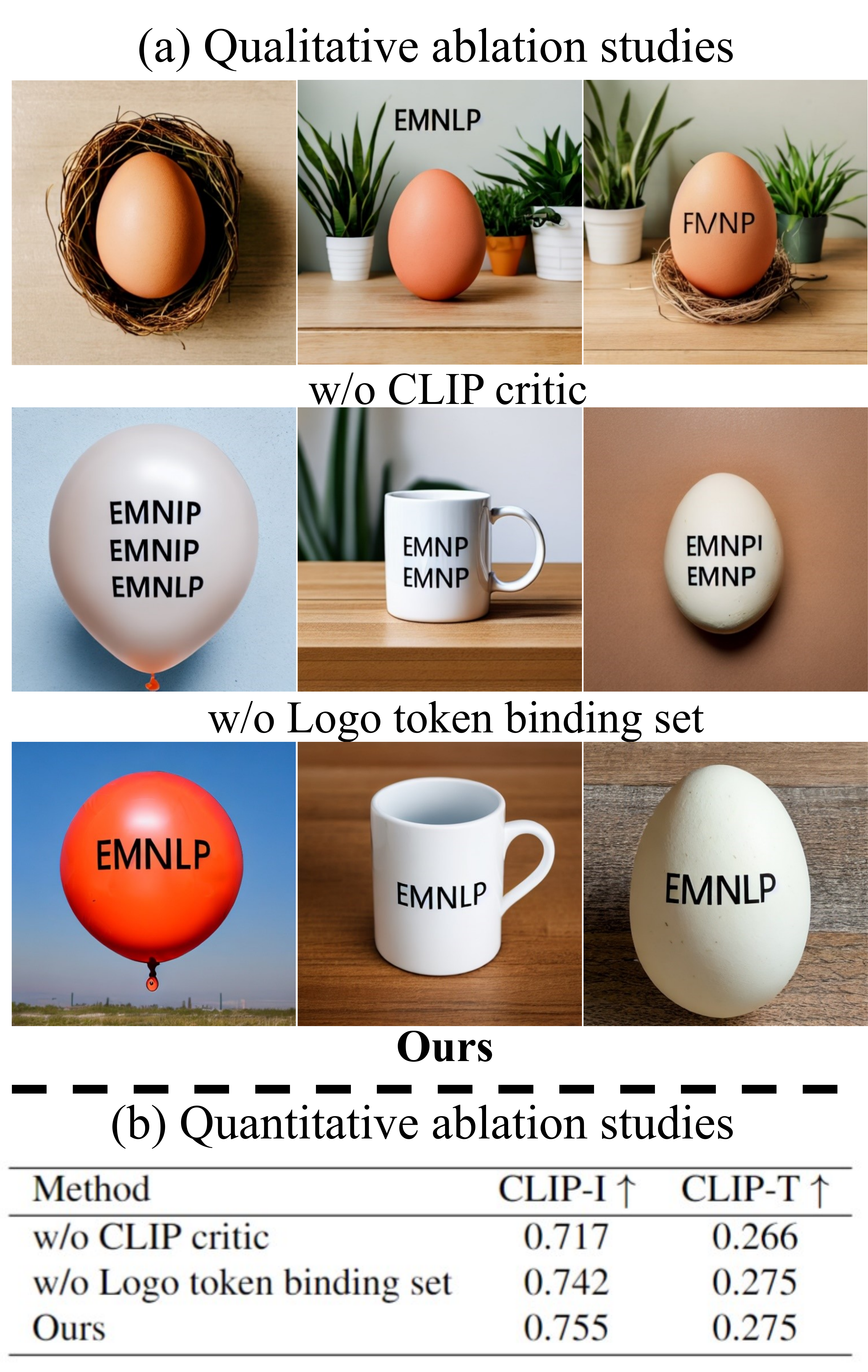}
},{\label{ablation_failure}} \textbf{Ablation study.} {(a) Qualitative (b) Quantitative ablation studies on the use of: (1) the CLIP critic for relation pre-training; (2) the logo token binding set for logo token binding.}]
\noindent Next, we conduct ablation studies to examine the effect of each  component of our LogoSticker qualitatively and quantitatively. We first ablate the effect of the CLIP critic during our relation pre-training. As illustrated in row 1 of Fig. \ref{ablation_failure}(a) and Fig. \ref{ablation_failure}(b), painting logos onto some objects can be difficult and not adequately learned without the CLIP critic. Although the model learns the identity of the logo, it does not know how to paint the logo onto certain objects. Quantitatively, it is reflected as the decreases in prompt and logo fidelities. Qualitatively, when prompted to generate a logo on certain objects, the model might generate the logo somewhere else or synthesize a logo with an incomplete identity.

\indent We then ablate the effect of the logo token binding set. From row 2 of Fig.~\ref{ablation_failure}(a) and Fig. \ref{ablation_failure}(b), we can see that, without the logo token binding set, the model cannot precisely recognize the target logo and thus cannot extract the logo identity accurately. The  model would be confused and tend to generate multiple or incomplete representations of the logo.
\end{figwindow}

\section{Conclusion}
\label{sec:Conclusion}

In this work, we conduct a pioneering exploration and introduce a novel task of logo insertion, aiming % which aims 
to insert diverse and complex logos accurately into diffusion models and enable identity-preserving and coherent generation of these logos in various contexts. We propose an effective two-phase pipeline LogoSticker for tackling this task, consisting of the actor-critic relation pre-training algorithm followed by the decoupled identity learning algorithm.
LogoSticker exhibits superior fidelities and coherence when generating diverse logos in various contexts, showing that the customization method can effectively deal with challenging logos, which encompass multilingual textual elements and intricate patterns. We hope our approach can benefit real-world applications like advertising.

\noindent \textbf{Acknowledgement.} This work was supported in part by the Research Grants Council under the Areas of Excellence scheme grant AoE/E-601/22-R, the National Natural Science Foundation of China (No. 62201484), HKU Startup Fund, and HKU Seed Fund for Basic Research.

%\clearpage  % TODO REVIEW/FINAL: This \clearpage needs to be removed from both review and camera-ready versions.

\bibliographystyle{splncs04}
\bibliography{main}
\end{document}